\documentclass[runningheads]{llncs}
\usepackage{graphicx}
\usepackage{stmaryrd}
\usepackage{amsfonts}
\usepackage{xcolor}
\usepackage{hyperref}
\begin{document}
\title{Mental Illness Classification on Social Media Texts using Deep Learning and Transfer Learning}
\author{Iqra Ameer\inst{1}\orcidID{0000-0002-1134-9713} \and
Muhammad Arif\inst{1}\orcidID{0000-0001-06141-02047} \and
Grigori Sidorov\inst{1}\orcidID{0000-0003-3901-3522} \and
Helena Gómez-Adorno\inst{2}\orcidID{0000-0002-6966-9912} \and
Alexander Gelbukh\inst{1}\orcidID{0000-0001-7845-9039} \and
}
\authorrunning{I. Ameer et al.}

\institute{Instituto Politécnico Nacional, Centro de Investigación en Computación, Mexico City, Mexico\\
\email{\{iameer2019,mariff2021,sidorov,gelbukh\}@cic.ipn.mx} \and 
Instituto de Investigaciones en Matemáticas Aplicadas y en Sistemas, Universidad Nacional Autónoma de México, Mexico City, Mexico\\
\email{helena.gomez@iimas.unam.mx}
}
\maketitle              % typeset the header of the contribution
\begin{abstract}
Given the current social distance restrictions across the world, most individuals now use social media as their major medium of communication. Millions of people suffering from mental diseases have been isolated due to this, and they are unable to get help in person. They have become more reliant on online venues to express themselves and seek advice on dealing with their mental disorders. According to the World health organization (WHO), approximately 450 million people are affected. Mental illnesses, such as depression, anxiety, etc., are immensely common and have affected an individual’s physical health. Recently Artificial Intelligence (AI) methods have been presented to help mental health providers, including psychiatrists and psychologists, in decision-making based on patients’ authentic information (e.g., medical records, behavioral data, social media utilization, etc.). AI innovations have demonstrated predominant execution in numerous real-world applications broadening from computer vision to healthcare. This study analyzes unstructured user data on the Reddit platform and classifies five common mental illnesses: depression, anxiety, bipolar disorder, ADHD, and PTSD. We trained traditional machine learning, deep learning, and transfer learning multi-class models to detect mental disorders of individuals. This effort will benefit the public health system by automating the detection process and informing appropriate authorities about people who require emergency assistance.

\end{abstract}

\section{Introduction}
Mental illness could be a sort of health condition that changes a person’s intellect, feelings, or behavior (or all three) and has been appeared to affect an individual’s physical health \cite{world2001world,marcus2012depression}. Mental health issues including depression,
schizophrenia, attention-deficit hyperactivity disorder (ADHD), autism spectrum disorder (ASD), etc., are highly prevalent today, and it is estimated that around 450 million people worldwide suffer from such problems\cite{world2001world}.

To way better get the mental health conditions and provide care, early detection of mental health problems is a basic step.
Different from the diagnosis of other chronic conditions that depend on research facility tests and measurements, mental illnesses are regularly diagnosed based on an individual’s self-report to particular surveys planned for the detection of specific patterns of feelings or social interactions \cite{hamilton1967development}.

Amid these uncertain times when COVID-19 torments the world, many people have indicated clinical anxiety or depression. This could be due to lockdown, limited social activities, a higher unemployment rate, economic depression, and fatigue related to work. American Foundation for Suicide Anticipation reported that individuals encounter anxiety (53\%) and sadness (51\%) more regularly now as compared to the time before covid-19 widespread.

Within the past decade, social media has changed social interaction. Along with sharing data and news, individuals effectively communicate their day-to-day activities, experiences, hopes, emotions, etc., and generate tons of data online. This textual data gives information that can be utilized to design systems to predict people's mental health. Moreover, the current limited social interaction state has forced people to express their thoughts on social media. It gives people an open stage to share their opinions with others numerous times attempt to find assistance \cite{murarka2021classification}.

A previous study explored the application of Machine Learning (ML) techniques in mental health \cite{shatte2019machine}. They reviewed literature by grouping them into four main application domains, such as detection and diagnosis (ii) prognosis, treatment and support, (iii) public health applications, and (iv) research and clinical administration. Another study explored the emerging area of application of DL techniques in psychiatry. They focused on DL by embedding semantically interpretable computational models of brain dynamics or behavior into a statistical machine learning context \cite{durstewitz2019deep}.

This study uses reddit.com\footnote{\url{https://www.reddit.com/} Last visited: 23-01-2022} user data proposed by Murarka and Radhakrishnan \cite{murarka2021classification} to determine mental illnesses, see sample of dataset instances in Table \ref{tab:dataset-examples}.  We applied traditional machine learning, deep learning, and transfer learning approaches to automatically detect mental disorders in social media texts. Our extensive experiments demonstrate that machine learning, deep learning, and transfer learning techniques have the potential to complement clinical procedures in the prediction of mental health between two classes of individuals: those seeking help online and those who are unaware of their condition.

   \begin{table}[th!]
    \centering
    %  \tiny
    %  \setlength\tabcolsep{1pt}
    \caption{\textbf{Sample instances of Reddit corpus}}
    \setlength{\tabcolsep}{5pt}
        \begin{tabular}{|l|p{9cm}|l|}
            \hline
            \textbf{No.} & \centering\textbf{Reddit Post} & \textbf{Label}\\
            \hline
            1 &  all the ideas that normally disappear as soon as we reach for a writing device will be captured and started. imagine all the projects we will begin and never finish! & adhd\\ 
            2 & i know this is long and i don't know if a lot of people will read this but i really just want to help. i had 2 panic attacks over the end of february and first day of march. i went to the doctor and had my blood work  &  anxiety\\ 
            3 & for example, did you ever notice that you had manic, hypomanic, depressive, etc. episodes? did you ever notice that sometimes you were "sad" and other times you were "excessively happy"? i'm in a sticky &  bipolar\\ 
            4 & i just feel so trapped and i *have* to do something about it. i don't know where i'll go or what i'll do to get by. i just can't stay here any longer. & depression\\ 
            5 & synesthesia. what is synesthesia? according to google, "synesthesia is a condition in which one sense (for example, hearing) is simultaneously perceived as if by one or more additional senses such as sight. & none\\ 
            6 & this is probably going to incite a lot of disagreement, maybe even anger, but that's okay; i'm going to say it anyway. anyone else tired of being told that just talking about your problems will solve your ptsd? & ptsd\\ 
            \hline
        \end{tabular}
        \label{tab:dataset-examples}
    \end{table}

The rest of the paper is organized as follows: section \ref{sec:RW} describes the studies on mental illness in literature. Section \ref{sec:PDD} explain the problem and gives dataset insights. Section \ref{sec:methods} give details of methodology applied to detect mental disorders. Section \ref{sec:results} presents results and their analysis. \ref{sec:conclusion} concludes the paper with possible future work.

\section{Related Work} \label{sec:RW}

In recent years, people have been using social media to communicate and seek advice on mental health issues. This has motivated researchers to take the information and apply a variety of NLP and ML approaches to help individuals who may want assistance. Initially, many researchers have focused on Twitter text \cite{orabi2018deep,benton2017multi,coppersmith2015clpsych}, later on the focus has shifted on Reddit platform \cite{kim2020deep,gkotsis2017characterisation,benton2017multi,zirikly2019clpsych}.

A wide range of approaches has been applied to mental health text analysis, from traditional ML to advanced DP. According to \cite{coppersmith2015clpsych}, they employed character-level language models to see how probable a user with mental health concerns would create a series of characters. \cite{benton2017multi} determined different types of mental health disorders by applying neural MTL, regression, and multi-layer perceptron single-task learning (STL) models. \cite{abusaa2004machine} trained the SVMs to distinguish 200 text messages into two classes: "ADHD or not." The most crucial step was the elimination of the acronym ADHD from the messages before learning, and further information concerning attention disorders was removed from the texts. The goal was to see how well the SVMS learns when keywords and even semantically relevant material are unavailable.

Deep feedforward neural network has outperformed typical ML models in a variety of data mining tasks \cite{amjad2020bend,amjad2021threatening}, and it has been used in the study of clinical and genetic data to predict mental health disorders. To diagnose depression, \cite{orabi2018deep} used word embeddings in combination with a range of neural network models such as CNNs and RNNs. To conduct binary classification on mental health textual posts, \cite{gkotsis2017characterisation} used Feed Forward Neural Networks, CNNs, traditional machine learning such as SVMs, and Linear classifiers. \cite{sekulic2020adapting} detected depression, ADHD, anxiety, and other types of mental illnesses by training a binary classifier for each disease with Hierarchical Attention Networks. The most recent work on this was a CNN-based classification model \cite{kim2020deep}. In which, the team trained a separate binary classifier for each type of mental disorder to conduct the detection. Reference \cite{hu2021explainable} found the potential factors to influence a person’s mental health during the Covid-19 pandemic by applying machine learning classifiers such as Naive Bayes, Logistic Regression, Support Vector Machine, Decision Tree, Random Forest, and Gradient Boosting. They have also presented an analysis of the feature selection technique LIME. 

In today's research world, transfer learning is extremely important. By using several types of transformers, researchers attempt to acquire greater accuracy and performance in each research study. The authors of \cite{murarka2021classification} examined three approaches for identifying and diagnosing mental illness on Reddit, including LSTM, BERT, and RoBERTa. Among these three methods, RoBERTa outperformed. Reference \cite{dhanalaxmi2020detection} employed RoBERTa to categorize COVID-19-related informative tweets, and their method yielded the best results.

%===========================================================================
\section{Problem Description and Dataset} \label{sec:PDD}

Murarka et al. \cite{murarka2021classification} developed a benchmark multi-class dataset from Reddit social media platform for mental illnesses detection. 

\subsection{Mental Illness Problem}
This study handles the mental illness problem as a multi-class classification problem. A text post of Reddit platform is given, and task is to classify the post into one of the six following mental disorder classes:

\begin{itemize}
    
    \item \textbf{ADHD (Attention Deficit Hyperactivity Disorder)}: A brain illness that affects how you pay attention, sit still, and control your behavior (common in children)\footnote{\url{https://www.cdc.gov/ncbddd/adhd/facts.html} Last visited: 24-01-2022}.
    
    \item \textbf{Anxiety}: A feeling of uneasiness, fear, and dread\footnote{\url{https://medlineplus.gov/anxiety.html\#:~:text=Anxiety\%20is\%20a\%20feeling\%20of,before\%20making\%20an\%20important\%20decision} Last visited: 24-01-2022}.
    
    \item \textbf{Bipolar}: Extreme mood swings, including emotional highs and lows, are a symptom of a mental health issue \footnote{\url{https://www.mayoclinic.org/diseases-conditions/bipolar-disorder/symptoms-causes/syc-20355955} Last visited: 24-01-2022}.
    
    \item \textbf{Depression}: A widespread and significant medical condition that has a negative impact on how someone feels, thinks, and acts\footnote{\url{https://www.psychiatry.org/patients-families/depression/what-is-depression} Last visited: 24-01-2022}.
      
    \item \textbf{PTSD (Post-traumatic stress disorder)}: A disorder that affects certain people after they have been through a traumatic, frightening, or dangerous incident\footnote{\url{https://www.nimh.nih.gov/health/topics/post-traumatic-stress-disorder-ptsd\#:~:text=Post\%2Dtraumatic\%20stress\%20disorder\%20(PTSD,danger\%20or\%20to\%20avoid\%20it.} Last visited: 24-01-2022}.
    
    \item \textbf{None}: No mental illness.
\end{itemize} 

\subsection{Dataset}
Reddit's post dataset is developed to
detect mental disorders into one of the five classes. The dataset
comprises a total of 16,930 posts. The posts were further divided into train, dev, and test groups with 13,726 posts in the training set, 1,716 posts in the dev set, 1,488 posts in the test set. Table \ref{tab:data_states} presents the number of posts for each mental illness class.

    \begin{table}[th!]
    \centering
    %  \small
    %  \setlength\tabcolsep{2pt}
    \caption{Number of posts for each mental illness class in whole dataset}
        \begin{tabular}{|c|c|}
            \hline
            ADHA & $3,023$\\
            Anxiety & $2,973$\\
            Bipolar & $2,956$\\
            Depression & $3.004$\\
            PTSD & $2,499$\\
            None & $2,478$\\
            \hline
            \textbf{Total} & $16,930$\\
            \hline
        \end{tabular}
        \label{tab:data_states}
    \end{table}
    
\begin{figure}[th!]
  \centering    
  \includegraphics[height=55mm]{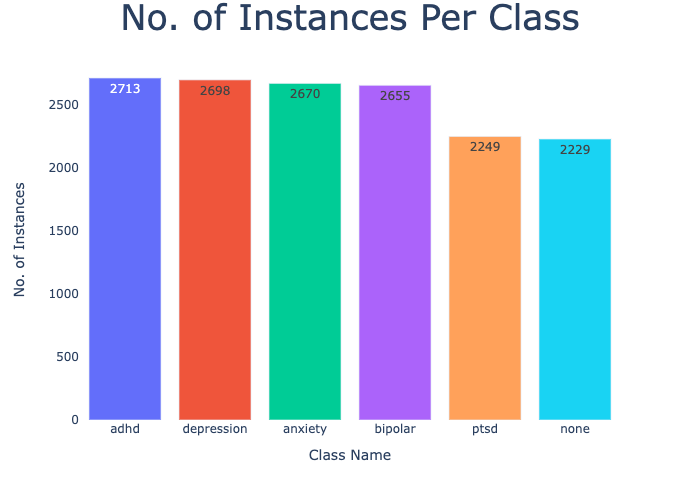}
  \caption{Mental Illness Dataset Statistics}
  \label{fig:data-stats}
\end{figure}

The dataset was already pre-processed by eliminating URLs or usernames containing sensitive material. However, we lowered-cased the post text,
removed the punctuation marks, removed stop words, and normalized the elongated words \cite{ameer2019cic,siddiqui2019bots,ameer2021author}, these are the most used pre-processing techniques in classification tasks. Figure \ref{fig:data-stats} shows the number of instances of training and test sets according to classes.

%===========================================================================

\section{Methods for Mental Illness Classification} \label{sec:methods}

This section describes our machine learning, deep learning, and transfer learning models applied for multi-class mental illness problems.

\subsection{Machine Learning Classifiers}

ML points at creating computational algorithms or statistical models that can consequently gather hidden patterns from the data \cite{pervaz2015identification,sittar2018multi}. For a long time, has seen an increasing number of ML models being created to analyze healthcare data \cite{murphy2012machine,biship2007pattern}. Conventional ML approaches require a significant sum of feature engineering for ideal performance--an essential step for most application scenarios to get excellent performance-- and time \cite{dwyer2018machine}.
Words help to create contextual content. Their sequence and structure can give important insights to classify texts \cite{ameer2019author,ameer2020multi}. In earlier studies, several researchers extracted word n-grams to classify user content on social media. \cite{kim2020deep} used word n-grams to detect mental illness from Reddit posts. Another study \cite{ive2020generation} utilized word n-grams to generate and evaluate artificial mental health records for NLP. We applied four different machine learning classifiers: Random Forest, Linear Support Vector Machine, Multinomial Naive Bayes, and Logistic Regression. 

The maximum number of features for each experiment was 1,000, i.e., we used the n-grams with the highest TF-IDF values. For the combination of word n-grams, the length of $N$ was minimum = 1 and maximum = 3 (We also tried $N$ = 1 and 2, and the results were not improved.

\subsection{Deep Learning Methods}

The second type of method is deep learning-based, in which different state-of-the-art neural network models were applied. Series of CLPsych shared task\footnote{\url{https://clpsych.org/} Last visited: 24-01-2022} an important role in development of mental health detection. We noticed that the most widely used models were Convolutional Neural Network, Recurrent Neural Networks, Long short-term memory, and Bidirectional Long short-term memory. 

In \cite{murarka2021classification}, authors applied Long short-term memory (LSTM) to detect mental illness from Reddit posts and achieved promising results. In \cite{mathur2020utilizing}, applied LSTM with attention mechanism to estimate suicidal intent by utilizing temporal psycholinguistic. We applied several pre-trained deep learning algorithms for multi-class mental illness detection such as Convolutional Neural Network, Gated recurrent unit (GRU), Bidirectional Gated recurrent units (Bi-GRU), LSTM, and Bidirectional LSTM.

We used Scikit-learn implementation of deep learning models considering the following parameters, which are usually the default: hidden layers = 3, hidden units = 64, no. of epochs = 10, batch size = 64, and dropout = 0.001. The parameters of CNN model are as follows: activation function = Rectified Linear Units (ReLU), optimizer= adam, hidden layers = 3, loss function = sigmoid, no. of epochs = 10, batch size = 64, dropout = 0.001.

\subsection{Transfer Learning Methods}

Bidirectional Encoder Representations from Transformers (BERT) is one of the foremost well-known advanced methods for NLP problems. RoBERTa (Robustly Optimized BERT Pretraining Approach) is another state-of-the-art language model that builds on BERT by modifying key hyperparameters and training on more data. It outperforms BERT on several benchmark tasks and forms the core of our proposed solution. The BERT model provided state-of-the-art performance over different NLP tasks without any critical task-specific design changes \cite{li2020enhancing,devlin2018bert}. \cite{9682721} used the BERT model for classification task on the multi-label text, which is trained by Google \cite{devlin2018bert}. The model was developed to enable transfer learning, which is why it went through a pre-training procedure that included utilizing both the BookCorpus and the English Wikipedia to help the model learn English. This training procedure consumes a lot of resources and time; therefore, fine-tuning a pre-trained model to a specific downstream is more efficient.

The pre-trained uncased version of the BERT base model was used in this study, which means that the text was converted to lowercase before the word tokenization stage. Each of the 12 encoders in the BERT base model has eight levels: four multi-head self-attention layers and four feed-forward layers.

In this study, the pre-trained XLNet was also used. The XLNet base model design consists of 12 transformer levels with 768 hidden layers and 12 attention head layers. To tokenize the sequences, the XLNet tokenizer was utilized. The tokens were then padded, and categorization was completed.

RoBERTa (Robustly Optimized BERT Pretraining Approach) is a cutting-edge language model that improves on BERT by tweaking key hyperparameters and training on additional data. It outperforms BERT on a number of benchmark tasks and serves as the foundation for our suggested solution. We applied a pre-trained RoBERTa base model. The RoBERTa base model design consists of 12 transformer levels with 768 hidden layers and 12 attention head layers. To tokenize the sequences, the RoBERTa tokenizer was utilized. The tokens were then padded, and categorization was completed.

%===========================================================================

\section{Results and Analysis} \label{sec:results}

    Our machine learning, deep learning, and transfer learning results are documented in Table \ref{tab:results}. In this Table, ``ML Algorithms" indicates traditional machine learning algorithms. The ``LinearSVC" indicates to Linear Support Vector Classifier, ``LR" indicates to Logistic Regression, ``NB" indicates to Naive Bayes, ``RF" indicates to Random Forest classifier. The ``DL Algorithms" indicates deep learning algorithms used in this study such as GNN, Bi-GNN, CNN, LSTM, and Bi-LSTM. The ``TL Algorithms" refers to pre-trained transfer learning algorithms applied to evaluate Reddit corpus, i.e., BERT, XLNet, and RoBERTa.
    
    Our pre-trained transfer learning RoBERTa model outperformed other traditional machine learning and deep learning algorithms with an accuracy score of 0.80, which is quite good on this challenging multi-class mental illness detection problem. The performance of the XLNet model is close to the RoBERTa difference of 0.01.
    
    The overall best results using the deep learning algorithm were on Bi-LSTM. This shows that Bi-LSTM is the most suitable algorithm to detect mental illness among deep learning algorithms. Interestingly, Bi-LSTM results are similar to the ones obtained with BERT. This indicates that the BERT model performs equally well on multi-class mental illness detection problems as advanced pre-trained transfer learning models. The accuracy scores of GRU, Bi-GRU, and CNN are not very high, highlighting the fact that multi-class mental illness detection on Reddit post's text is a challenging task. 
    
    Using traditional machine learning algorithms, overall, best results are obtained using combination of word n-grams when length of \textit{N} was minimum = 1, maximum = 3 (Accuracy = 0.78, F1 = 0.67). This shows that combinations of word grams (length of \textit{N} =  1-3) were the most suitable features when we trained the model on the Reddit social media platform. 
    
    The RoBERTa model's detailed class-wise results are shown in Table \ref{tab:class-wise-results}. The first surprising outcome is the model's ability to recognize non-illness-related postings with high accuracy. The model can categorize the \texttt{none} class with an F1 score of more than 0.98. That gives us the indication that when it comes to detecting mental illness on social media, this model will suffer from very few false positives.
   
    The two highest performing classes among mental disorder are \texttt{adhd} and \texttt{ptsd}, while on the contrary the two poorest performing classes are \texttt{depression} and \texttt{anxiety}.

There are number of factors contribute to the result of \texttt{depression} and \texttt{anxiety} classes. \texttt{depression} and \texttt{anxiety} have the fewest average number of tokens (words) per post among all classes. 
    
       \begin{table}[ht!]
     \caption{Results obtained by applying various classical machine learning, deep earning and transfer learning techniques}
    \centering
    %  \tiny
    %  \setlength\tabcolsep{2pt}
        \begin{tabular}{|c|c|c|}
            \hline
             \textbf{ML Algorithms} &  \textbf{Accuracy} & \textbf{F1-score}\\
            \hline
            \multicolumn{3}{|c|}{\textbf{Classical Machine Learning}}\\
            \hline 
            LinearSVC & $0.79$ & $0.80$\\
            LR & $0.79$ & $0.80$\\
            NB & $0.74$ & $0.75$\\
            RF & $0.75$ & $0.76$\\
            \hline
            \multicolumn{3}{|c|}{\textbf{Deep Learning}}\\
            \hline
            \textbf{DL Algorithm} &  \textbf{Accuracy} & \textbf{F1-score}\\
            \hline
            GRU & $0.62$ & $0.64$\\
         
            Bi-GRU & $0.63$ & $0.65$\\
       
            CNN & $0.64$ & $0.65$\\
       
            LSTM & $0.76$ & $0.77$\\
   
            Bi-LSTM & $0.78$ & $0.79$\\
            \hline
            \multicolumn{3}{|c|}{\textbf{Transfer Learning}}\\
            \hline
            \textbf{DL Algorithm} &  \textbf{Accuracy} & \textbf{F1-score}\\
            \hline
            BERT& $0.78$ & $0.80$\\
            XLNet& $0.79$ & $0.80$\\
            RoBERTa& $0.83$ & $0.83$\\
            \hline
        \end{tabular}
        \label{tab:results}   
    \end{table}

%===========================================================================
% OLD RESULTS
    %   \begin{table}[ht!]
    %  \caption{RoBERTa class-wise results}
    % \centering
    % %  \tiny
    % %  \setlength\tabcolsep{2pt}
    %     \begin{tabular}{|c|c|c|c|}
    %         \hline
    %          \textbf{Class} &  \textbf{Precision} & \textbf{Recall} & \textbf{F1-score}\\
    %         \hline 
    %         adha & $0.77$ & $0.78$ & $0.77$\\
    %         anxiety & $0.67$ & $0.78$ & $0.72$\\
    %         bipolar & $0.76$ & $0.76$ & $0.76$\\
    %         depression & $0.77$ & $0.66$ & $0.66$\\
    %         none & $0.92$ & $0.94$ & $0.94$\\
    %         ptsd & $0.88$ & $0.82$ & $0.82$\\
    %         \hline
    %     \end{tabular}
    %     \label{tab:class-wise-results}   
    % \end{table}

%===========================================================================

For example, when comparing \texttt{depression} and \texttt{ptsd} classes, \texttt{depression} class have around 53\% fewer textual content. Furthermore, researchers demonstrate that \texttt{depression} is frequently associated with another mental disorder, and our findings support this. In  12\% percent of \texttt{ptsd} posts, 12\% percent of \texttt{anxiety} posts, and 31\% of \texttt{bipolar} posts, the term depression appears. Anxiety is present in 12\% of \texttt{adhd} posts, 14\% of \texttt{bipolar} posts, and 20\% of \texttt{ptsd} posts, respectively. This means that, unlike with the rest of the diseases, the model is unable to place a high value on the mention of these class names, making the categorization of these two labels challenging.

 %Confusion matrix   
    
    \begin{figure}[th!]
  \centering    
  \includegraphics[scale=0.45]{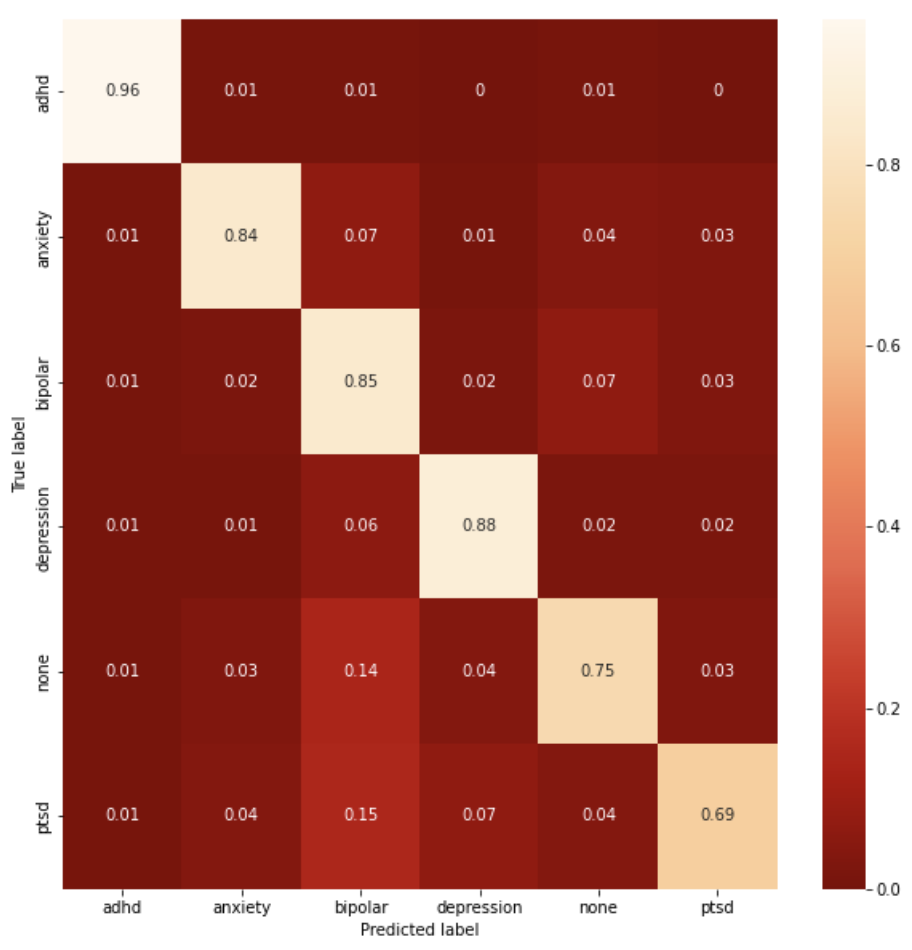}
  \caption{RoBERTa confusion matrix}
  \label{fig:cm}
\end{figure}

       \begin{table}[ht!]
     \caption{RoBERTa class-wise results}
    \centering
    %  \tiny
    %  \setlength\tabcolsep{2pt}
        \begin{tabular}{|c|c|c|c|}
            \hline
             \textbf{Class} &  \textbf{Precision} & \textbf{Recall} & \textbf{F1-score}\\
            \hline 
            adha & $0.85$ & $0.83$ & $0.84$\\
            anxiety & $0.73$ & $0.78$ & $0.76$\\
            bipolar & $0.83$ & $0.76$ & $0.80$\\
            depression & $0.76$ & $0.83$ & $0.70$\\
            ptsd & $0.90$ & $0.87$ & $0.88$\\
            none & $0.99$ & $0.96$ & $0.98$\\
            \hline
        \end{tabular}
        \label{tab:class-wise-results}   
    \end{table}
    
Figure \ref{fig:cm} shows the confusion matrix for RoBERTa model. The terms \texttt{depression} and \texttt{anxiety} are mentioned in more instances in the \texttt{adhd} and \texttt{ptsd} classes than the name these class themselves. One could expect poor outcomes as a result of this, but these classes outperform all others. This exhibits the actual potential of our approach since it does not depend solely on the mention of class names in the post but also has a deep awareness of the post's context.
    
\section{Conclusion} \label{sec:conclusion}

The present Covid-19 outbreak and globally forced isolation are our primary motivations for multi-class mental illness detection efforts.
We feel that social media platforms have become the most widely used communication medium for individuals, allowing them to express themselves without fear of being judged. We applied state-of-the-art traditional machine learning, deep learning, and transfer learning-based methods for multi-class mental illness detection problem. The best results (see Table \ref{tab:results}) obtained using pre-trained RoBERTa transfer learning model (accuracy = 0.83, F1-score = 0.83).

In the future, we plan to develop a multi-label dataset for mental illness problems, which would be more reflective of the situation than a multi-class dataset, as a post can have more than one mental disease instead of one per post, i.e., depression, anxiety. We can also use the data augmentation technique on top of this existing mental health data \cite{amjad2020data}. We plan to apply other transfer learning-based models such as DistilBERT, etc., in the future. An ensemble modeling would be considered to improve classification performance.

 \section*{Acknowledgements}
% %
The work was done with support from the Mexican Government through the grant A1-S-47854 of the CONACYT, Mexico and grants 20211784, 20211884, 20211178 of the Secretaría de Investigación y Posgrado of the Instituto Politécnico Nacional, Mexico, and grants of PAPIIT-UNAM project TA101722. The authors utilize the computing resources brought to them by the CONACYT through the Plataforma de Aprendizaje Profundo para Tecnologías del Lenguaje of the Laboratorio de Supercómputo of the INAOE, Mexico.

\bibliographystyle{splncs04}
\bibliography{Mental-illness-v2}




% LNCS Homepage, \url{http://www.springer.com/lncs}. Last accessed 4
% \end{thebibliography}
\end{document}